%% file: competitionBootstrap.tex
\documentclass[a4paper]{article}

\usepackage{geometry}
\usepackage{authblk}
\usepackage{natbib}
\usepackage{times}
\usepackage{helvet}
\usepackage{courier}
\usepackage{longtable}
\usepackage{subfigure}
\usepackage{graphicx}
\usepackage{bm}
\usepackage{booktabs}
\usepackage[usenames,dvipsnames,table]{xcolor}
\usepackage{amsmath,amsfonts}

\include{macros}

\begin{document}

\date{9 August 2023}

\renewcommand\Authsep{, }
\renewcommand\Authand{, }
\renewcommand\Authands{, }

\title{Competitions in AI -- Robustly Ranking Solvers \\ Using Statistical Resampling}
\author[1]{Chris Fawcett}
\author[2]{Mauro Vallati}
\author[1,3,4]{Holger H. Hoos}
\author[5]{Alfonso E. Gerevini}
\affil[1]{Department of Computer Science, University of British Columbia, Canada}
\affil[2]{School of Computing and Engineering, University of Huddersfield, United Kingdom}
\affil[3]{Chair for AI Methodology, RWTH Aachen University, Germany}
\affil[4]{LIACS, Leiden University, The Netherlands}
\affil[5]{Department of Information Engineering, University of Brescia, Italy}


\maketitle

\begin{abstract}

Solver competitions play a prominent role in assessing and advancing the state of the art for solving many problems in AI and beyond.
Notably, in many areas of AI, competitions have had substantial impact in guiding research and
applications for many years, and for a solver to be ranked highly in a competition carries considerable weight.
But to which extent can we expect competition results to generalise to sets of problem instances different from those used in a particular competition?
This is the question we investigate here, using statistical resampling techniques.
We show that the rankings resulting from the standard interpretation of competition results can be very sensitive to even
minor changes in the benchmark instance set used as the basis for assessment and can therefore not be expected to carry over to other samples from the same underlying instance distribution.
To address this problem, we introduce a novel approach to statistically meaningful analysis of competition results based on resampling performance data.
Our approach produces confidence intervals of competition scores as well as statistically robust solver rankings with bounded error.
Applied to recent SAT, AI planning and computer vision competitions, our analysis reveals frequent statistical ties in solver performance as
well as some inversions of ranks compared to the official results based on simple scoring.

\end{abstract}

\section{Introduction}


Competitions in AI play a prominent role in assessing and improving the state of the art,
by consistently evaluating a set of solvers at a fixed point in time.
Particularly well-known examples include the SAT and AI planning competitions that have been running for nearly two decades,
and the ImageNet Large Scale Visual Recognition Challenge (ILSVRC) which has been run annually since 2010
(see, e.g., \citeauthor{simonEtAl2005satcomp} \citeyear{simonEtAl2005satcomp}; \citeauthor{belov2014sat} \citeyear{belov2014sat}; \citeauthor{mcdermott20001998} \citeyear{mcdermott20001998}; \citeauthor{survey2014} \citeyear{survey2014}; \citeauthor{russakovsky2015imagenet} \citeyear{russakovsky2015imagenet}).
The outcomes of such competitions guide research as well as applications, and they often have remarkable impact on their respective research areas. In particular, top-ranking solvers enjoy great visibility and popularity in the community and beyond.




It is known that competition results can be significantly affected by changing the set of benchmark instances used, the way in which instances are described~\cite{vallati2015effective} and
the execution environments used for evaluation~\cite{howe2002critical}. Moreover, final ranks have been shown 
to also depend on the pseudo-random number seeds used in randomised solvers~\cite{hurleystatistical} as well as the settings of solver parameters~\cite{HutEtAl17}.


In this work, we focus on the impact of the benchmark instances on competition results
-- an important issue that is largely orthogonal to other factors. 
The construction of benchmark sets for use in competitions is a complex task, and a number of recent publications
have focused on various aspects involved~(see, e.g., \citeauthor{hoos2013bench} \citeyear{hoos2013bench}; \citeauthor{hoos2012homo} \citeyear{hoos2012homo}; \citeauthor{survey2014} \citeyear{survey2014}; \citeauthor{vallati2015towards} \citeyear{vallati2015towards}).



Realistically, we cannot expect competition results -- in the form of solver rankings -- to remain the same
when considering benchmark sets markedly different from those used in a competition.
One reason for this lies in the fact that in mature areas of AI, 
it is rare to
achieve progress that enables a new solver to dominate the previous state of the art on all types of instances,
and the state of the art in solving the respective general problems is often comprised of a set (or portfolio) of solvers 
with complementary strengths
(see, e.g., \citeauthor{satzilla} \citeyear{satzilla}; \citeauthor{pbp} \citeyear{pbp}; \citeauthor{ibacop} \citeyear{ibacop}; \citeauthor{Seipp2015} \citeyear{Seipp2015}).


However, one would hope that competition scores and solver rankings are relatively unaffected at least by 
small changes to the given benchmark instance set.
Surprisingly, this often not the case, as illustrated by a simple experiment, in which for each track of the 
recent SAT, AI planning and ILSVRC competitions considered later in this work,
we measured the effect of removing a single problem instance (or image category for ILSVRC) on the official competition rankings.
As seen in Table~\ref{tab:singleInstanceRemovalSummary},
in almost all tracks, there were multiple problem instances whose removal caused changes in the competition rankings.
For the SAT and AI planning competition tracks, many of these single-instance removals caused changes in the composition or order of the top 10 and even top 3 solvers.

More generally, for competition results to be maximally informative outside of the competition setting,
scores and rankings should ideally be stable across samples from the distribution $\mathcal{D}$ underlying the competition instance set.
We note that, conceptually, this distribution exists even if hand-crafted or specific real-world instances are used;
clearly, generating many such samples and performing evaluations on all of them can be costly or infeasible.

In this work, we investigate the degree to which competition scores and rankings are stable under an
approximation of a large set of samples from $\mathcal{D}$.
The key idea is to resample a given set of benchmark instances -- and the corresponding competition results -- in order to
approximate drawing new samples from the underlying distribution $\mathcal{D}$. 
For each sample thus generated, solver scores and rankings are determined,
and these are used to estimate confidence intervals for solver competition scores as well as the stability of solver rankings.
Furthermore, we use these samples to generate a more robust competition ranking, grouping solvers without statistically significant performance differences,
with bounded family-wise error, using bootstrap hypothesis testing with multiple-testing correction.




In planning competitions, statistical analyses have often been used post-competition to evaluate solver performance, and more rarely to generate rankings
(see, e.g., \citeauthor{longFox2003satcomp} \citeyear{longFox2003satcomp}; \citeauthor{gerevini2009deterministic} \citeyear{gerevini2009deterministic}; \citeauthor{ipc2011} \citeyear{ipc2011}).
Our work expands these approaches in several ways; in particular:
(i) we consider resampled sets of benchmark instances, allowing for a robust performance assessment across an approximation of the underlying instance distribution $\mathcal{D}$;
(ii) we derive confidence intervals for competition scores over samples from $\mathcal{D}$;
(iii) we bound the type I error across multiple statistical tests between pairs of solvers.


Our results show that in the 17 tracks from 3 recent SAT, AI planning and ILSVRC competitions we analysed,
the ranks and scores of many solvers are not stable even under small changes to the benchmark instance set,
that can be expected to occur between samples from the same underlying distribution,
and many solvers are in fact statistically tied when these small changes are considered.
We found that, for example, in the main track of the 2016 SAT Competition, 17 solvers are tied for first place in our robust competition rankings.
We also observed several cases where official solver ranks are inconsistent with our robust ranking in the SAT and ILSVRC competition data.




\begin{table}
\footnotesize
    \begin{center}
        \begin{scriptsize}
        \begin{tabular}{rr@{\hskip 2em}rrrrr}
            \input{tables/single_instance_removal_summary}
        \end{tabular}
        \end{scriptsize}
        \caption{%
            For each of the SAT 2016, IPC 2014 and ILSVRC 2015 competition tracks considered later,
            we show the number of problem instances from the resp.~competition benchmark set that, when individually removed from that set, lead to solver rankings different from the official competition results.
            We also present separate counts for any changes in ranking (all), 
            changes in the composition of the set of top solvers (comp) and 
            ranking of the top solvers (order).
            \vspace{-0.4cm}
        }
        \label{tab:singleInstanceRemovalSummary}
    \end{center}
\end{table}

\section{Bootstrap analysis of competition data}

Given a set of solvers $\mathcal{S}$, a set of benchmark problem instances $\mathcal{I}$, and a set $\mathcal{R} \subseteq \mathcal{I} \times \mathbb{N}$ of \emph{runs}, each of which is specified by an instance and a pseudo-random number seed,
official competition data typically consists of the results of executing each $s \in \mathcal{S}$ for each $\left(i, \mathsf{seed}\right) \in \mathcal{R}$.
Each of these run results is composed of run success/failure information (including crashed runs or
those that produce incorrect output), the solution or solution quality obtained by the solver, and the resources used
(e.g., CPU time).
This run data is then aggregated for each solver by a (competition-specific) scoring mechanism, and
solvers are ranked by score to produce the official competition results.

Introduced in the seminal paper by Efron~(\citeyear{efron1979bootstrap}), bootstrap methods are
a family of resampling approaches that, given an existing sample $S$ from an assumed
underlying distribution, generate a set of additional samples by re-sampling $S$ uniformly
with replacement. This new set of so-called bootstrap samples can be used to estimate statistics of
the underlying distribution, to generate confidence intervals for statistics calculated
using the original sample and for hypothesis testing (see, e.g., \citeauthor{davisonHinkley1997bootstrap} \citeyear{davisonHinkley1997bootstrap};
\citeauthor{efronTibshirani1986bootstrap} \citeyear{efronTibshirani1986bootstrap}).

In the following, we generate bootstrap samples from a given set of competition data,
by sampling uniformly at random from $\mathcal{R}$, with replacement, to create an alternative set of runs $\mathcal{R}^\prime$ of the same size as $\mathcal{R}$.
We then select the results for the runs in $\mathcal{R}^\prime$ (these are all contained in the original competition data for the runs in $\mathcal{R}$), and use the competition scoring mechanism to determine $\mathsf{score}\left(s, \mathcal{R}^\prime\right)$, the score of solver $s$ on benchmark set $\mathcal{R'}$, for each $s \in \mathcal{S}$.
We thus obtain a set of \emph{bootstrap replicates} of the competition.
From these replicates, we can determine the $\left(1-\alpha\right)$ confidence interval $\left[s_l, s_h\right]$ for the score of each solver $s$, using the percentile method, where $s_l$ and $s_h$ are the $\alpha/2$ and $\left(1-\alpha/2\right)$ quantiles of the distribution of scores for $s$ over the bootstrap samples, respectively.
Consistent with common practice for bootstrap analyses, we produce 10\,000 bootstrap samples for each of the considered competition tracks and use $\alpha:=0.05$.

\subsection{Bootstrap hypothesis testing}

Given $k$ bootstrap replicates of a competition and two solvers $s_1$ and $s_2$, we wish to design a one-sided significance test
for the null hypothesis $H_0 : \mathsf{score}\left(s_1, \mathcal{D}\right) \leq \mathsf{score}\left(s_2, \mathcal{D}\right)$,
where $\mathsf{score}\left(s, \mathcal{D}\right)$ is the score for solver $s$ aggregated over the underlying problem instance (and therefore run)
distribution $\mathcal{D}$. In other words, our null hypothesis is that the performance of solver $s_1$ is equal to or worse than that of $s_2$ over $\mathcal{D}$.

We now count the number of bootstrap replicates for which the score of $s_1$ is less than or equal to that of $s_2$. If, and only if, the ratio between this count and $k$, the total number of samples, is less than a given significance threshold $\alpha$, we reject $H_0$ with a $p$-value equal to that ratio.
Since our bootstrap samples were drawn from $\mathcal{R}$, which itself is a sample of $\mathcal{D}$, this yields the desired hypothesis test. 
We note that this bootstrap test implicitly exploits the duality between confidence intervals (in this case, for the difference between the scores of $s_1$ and $s_2$) and hypothesis tests and does not require any distributional assumptions for $\mathcal{D}$.
Consistent with common practice, we set $\alpha:=0.05$. 

\subsection{Robust ranking}

This hypothesis test, along with our set of bootstrap replicates, is then used to produce an alternative, robust ranking
of the solvers in $\mathcal{S}$. First, we consider the problem of identifying the solver (or set of solvers) that should win a given competition. A natural choice is to declare as the winner the $s^\ast \in S$ that achieve the highest score in the greatest number of our bootstrap samples, and which therefore has the highest empirical probability of winning the competition under our resampling scheme. (We note that ties could arise at this stage, leading to multiple $s$ being selected, which does not cause any complications in the steps that follow.)
We then determine which of the remaining solvers is statistically tied with the winner, by using our
pairwise bootstrap hypothesis test for $s^\ast$ and each remaining solver.%
\footnote{Robust rankings can be determined using arbitrary tests for pairwise performance differences.
	We have also explored the use of a permutation test for the same $H_0$ and obtained very similar results. We prefer the bootstrap test, since it can make use of the same set of bootstrap replicates as computation of confidence intervals and does not require adjustments to preserve statistical power in situations where $H_0$ does not hold.}

Since we are performing multiple hypothesis tests involving $s^\ast$, there will be a compounding probability
of false rejections (incorrectly assessing some solver to perform significantly worse than $s^\ast$).
We use the Holm-Bonferroni multiple-testing correction to ensure an upper bound of $\alpha:=0.05$ for the overall (family-wise) false rejection probability \cite{Holm1979simple}.
This multiple testing correction works as follows. We perform pairwise tests between the winner $s^\ast$ and the $m = \left|S\right|-1$ remaining solvers $s_1, s_2, ..., s_m$.
Let $H_1, H_2, ..., H_m$ be the null hypotheses for these tests, with corresponding $p$-values $p_1, p_2, ..., p_m$. 
We sort these $p$-values in ascending order $p^\prime_1, p^\prime_2, ..., p^\prime_m$ and let $i$ be the lowest index such that $p^\prime_i < \frac{\alpha}{m+1-i}$.
If $i=1$, we do not reject any of the null hypotheses, and if no such $i$ exists, we reject all null hypotheses.
In all other cases, we reject the null hypotheses $H^\prime_1, ..., H^\prime_{i-1}$, and we do not reject $H^\prime_i, ..., H^\prime_m$. The corresponding
solvers $s^\prime_i, ..., s^\prime_m$ are grouped together with $s^\ast$ as statistically tied winners.

After choosing the group of winners, we proceed iteratively by removing $s^\prime_i, ..., s^\prime_m$ and $s^\ast$ from consideration and repeating the process until zero or one solvers remain. The result of this process is an ordered ranking of \emph{groups} of statistically tied solvers.
We note that our procedure bounds the probability of making errors (by incorrectly ranking a solver lower than justified statistically) in any iteration at $\alpha$; therefore, it gives us statistical confidence in the grouping of solvers identified in each iteration.

\section{SAT competition analysis}

\begin{table*}
\footnotesize
    \begin{center}
        \begin{scriptsize}
        \begin{tabular}{r@{\hskip 1em}rr@{\hskip 1em}rrrr@{\hskip 1em}rrrr@{\hskip 1em}rrrr}
            \input{tables/competition_summaries}
        \end{tabular}
        \end{scriptsize}
        \caption{%
            Summary of the SAT 2016, IPC 2014 and ILSVRC 2015 competition tracks considered in our analysis.
            \vspace{-0.2cm}
        }
        \label{tab:competitionSummaries}
    \end{center}
\end{table*}

We first consider six tracks from the recent 2016 SAT Competition\footnote{http://baldur.iti.kit.edu/sat-competition-2016}. These tracks
are \emph{main, main-application, main-crafted, agile, random} and \emph{nolimit}.
Table~\ref{tab:competitionSummaries} shows the number of competing solvers and competition benchmark instances
for each of the tracks. In all cases, we generated 10\,000 bootstrap samples from the competition data.

Due to limited space, we focus our discussion on the results for the \emph{main} track, but will briefly discuss results for other tracks. Complete
results for all tracks are available in our Supplementary Material. 

Table~\ref{tab:empiricalFirstPlace} shows the fraction of our bootstrap samples for which solvers in the 
\emph{main} track were ranked first. Incredibly, 16 of the 29 competing solvers were ranked first in at least one bootstrap sample.
\emph{MapleCOMSPS\_DRUP}, the official winner, was only ranked first in 38.16\% of our bootstrap samples.
Furthermore the solver ranked third in the competition, \emph{Lingeling bbc main default}, was ranked first in fewer samples
than the next 13 solvers combined.
This illustrates how little confidence we should have to obtain precisely the same winner when repeating the competition on a statistically indistinguishable instance set.

\begin{figure*}[!t]
    \footnotesize
    \begin{center}
        \subfigure[SAT 2016 \emph{main}]{
            \includegraphics[width=0.315\textwidth]{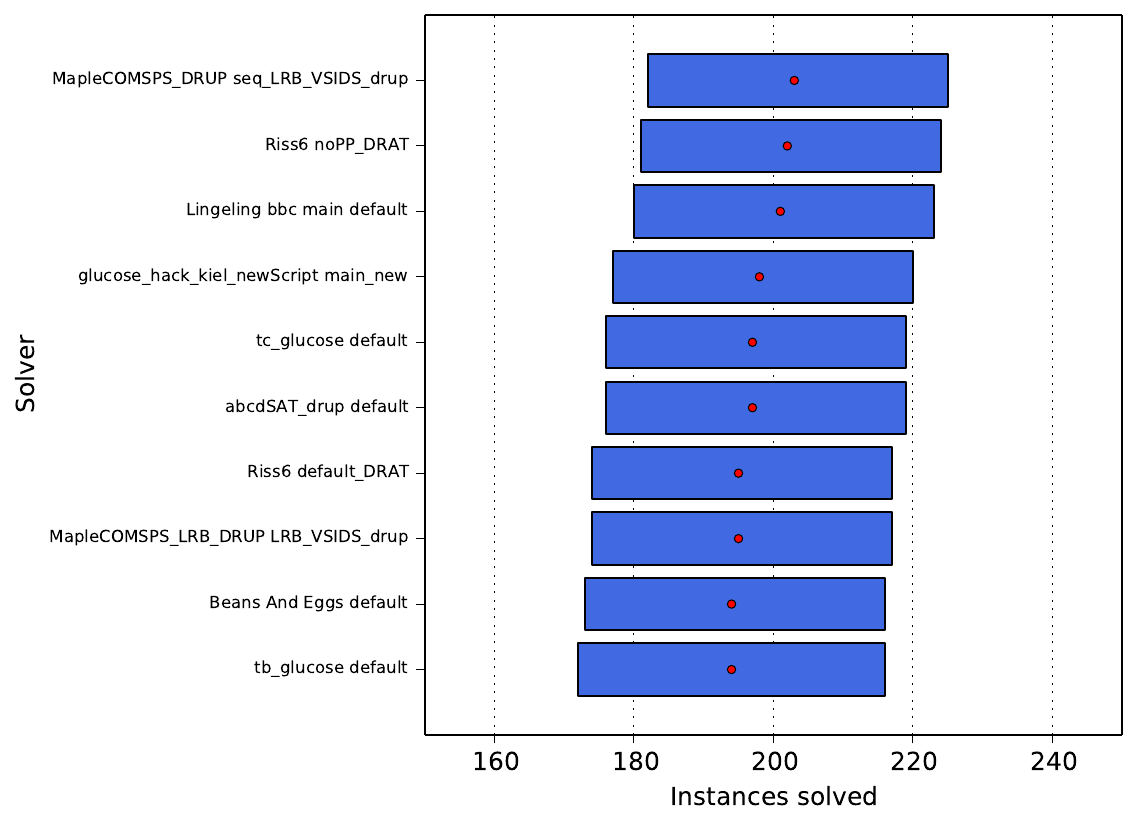}
            \label{fig:sat2016bootstrapCI}
        }
        \subfigure[IPC 2014 \emph{satisficing}]{
            \includegraphics[width=0.335\textwidth]{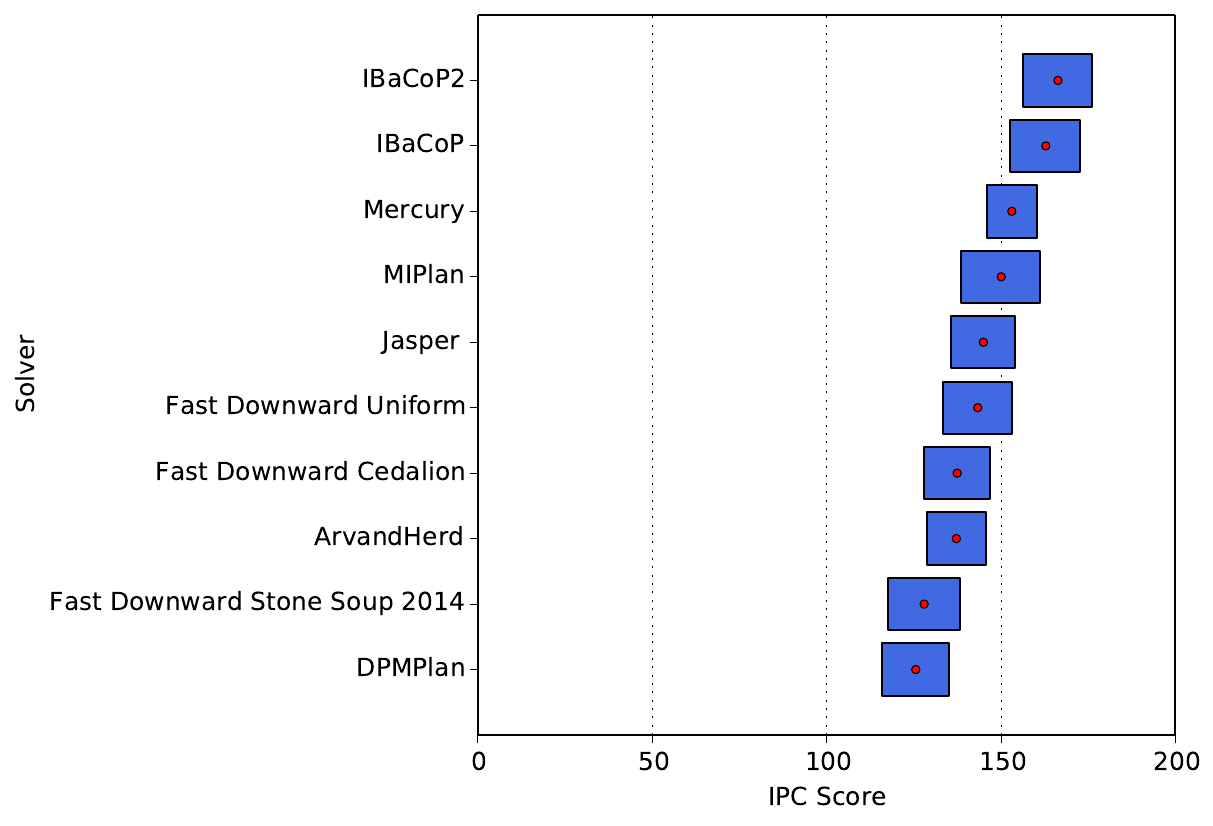}
            \label{fig:ipc2014bootstrapCI}
        }
        \subfigure[ILSVRC 2015 \emph{classification}]{
            \includegraphics[width=0.30\textwidth]{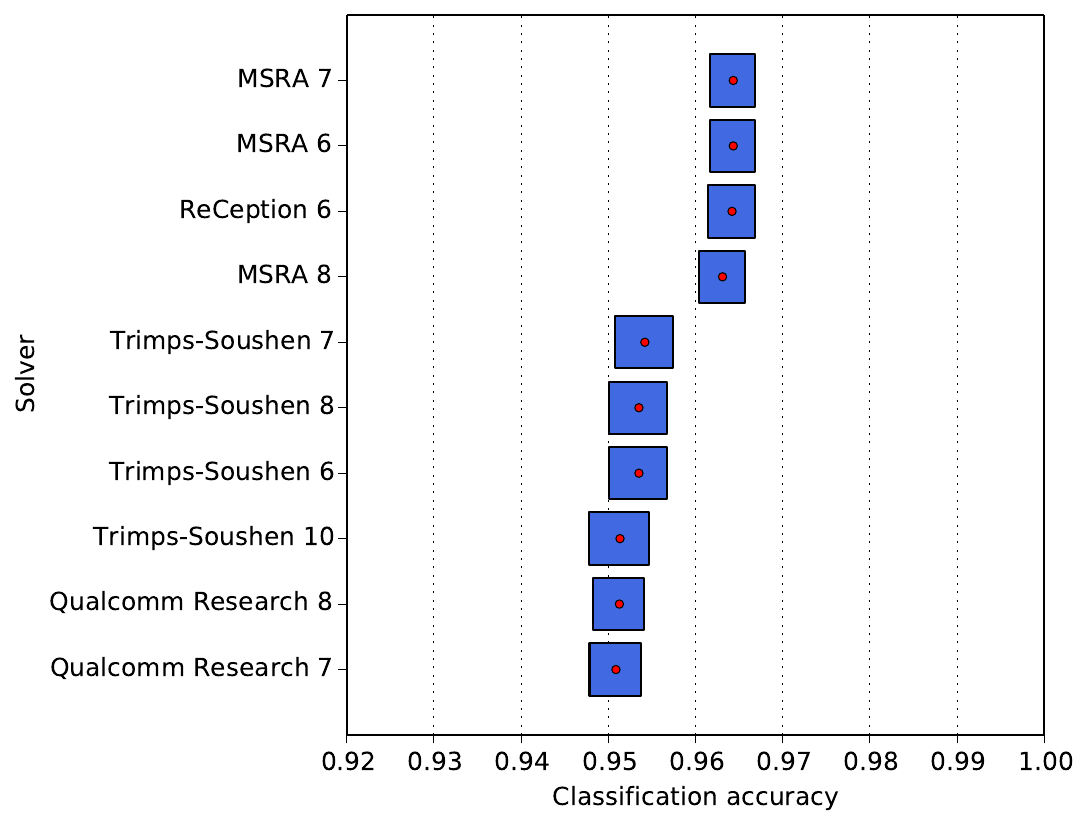}
            \label{fig:imagenet2015bootstrapCI}
        }
        \caption{%
            Using our 10\,000 bootstrap replicates for the SAT 2016 \emph{main} track, IPC 2014 \emph{satisficing} track
            and ILSVRC 2015 \emph{object classification+localization} track (classification accuracy),
            we plot the 0.95 confidence intervals of the respective competition scores for the top 10 solvers based on official competition rankings.
            The official competition score for each solver is indicated by a red point within the respective confidence interval.
            \vspace{-0.4cm}
        }
        \label{fig:bootstrapConfidenceIntervals}
    \end{center}
\end{figure*}

We present the 0.95 confidence intervals for competition scores of the top 10 solvers (based on official competition rankings) in Figure~\ref{fig:sat2016bootstrapCI}.
Along with the mean inter-quartile range shown in Table~\ref{tab:competitionSummaries}, this figure demonstrates that there is a large overlap in performance among the competitors.

\begin{table}
\footnotesize
    \begin{center}
        \begin{scriptsize}
        \begin{tabular}{r@{\hskip 1em}l}
            \input{tables/empirical-first-place-truncated}
        \end{tabular}
        \end{scriptsize}
        \caption{%
            Considering the solvers in the SAT 2016 \emph{main} track, IPC 2014 \emph{satisficing} track, and ILSVRC 2015 \emph{object classification+localization} track (classification accuracy),
            we count in how many of our 10\,000 bootstrap replicates that solver was ranked first, using the official competition ranking mechanism (including ties).
            In the case of SAT 2016, we show counts for the top-3 solvers along with an aggregate count over the remaining solvers ranked first in at least one sample.
            For the other two competitions, the solvers shown are the only ones ranked first in at least one sample.
            We note that the presence of ties means that the individual solver fractions can sum to more than 100\%.
            \vspace{-0.2cm}
        }
        \label{tab:empiricalFirstPlace}
    \end{center}
\end{table}

A robust ranking for this track, computed using the procedure described previously, is shown in Table~\ref{tab:exampleAltRankings}.
As we can see, 17 solvers show performance differences in the bootstrap replicates that should not be considered significant,
resulting in a 17-way tie for first place!
We ordered the solvers within each group shown in Table~\ref{tab:exampleAltRankings} by their median score across all bootstrap samples,
and number each group of statistically tied solvers consecutively.
We also show the fractional rank assigned to each solver in a group,
where fractional ranks are computed such that the overall sum of all ranks is the
same, regardless of ties.

Results for the other considered tracks showed a very similar trend, especially for the related \emph{main-application} and \emph{main-crafted} tracks.
One exception was the \emph{agile} track, where \emph{Riss6 default} was the single winning solver in our robust ranking, with a 4-way tie for second place.

Interestingly, we found two cases of rank inversions, where a solver $A$ was ranked below another solver $B$ in the official results,
but appeared in a group above that of $B$ in our robust rankings.
Specifically, in the \emph{main-crafted} track, \emph{MapleGlucose default} (competition rank 14) had a fractional rank of 7.0 (group 1), while \emph{Splatz 06v main default} (13) had fractional rank 17 (group 2) in our results.
Similarly, in the \emph{agile} track, \emph{glue\_alt default} (competition rank 6) had a fractional rank of 3.5 (group 2), while \emph{tc\_glucose\_agile default} (5) had a fractional rank of 6 (group 3).
While these inversions involve solvers outside of the official competition podium ranks, these situations could have just as easily occurred between highly-ranked solvers given slightly different competition data.

\begin{table}[t!]
\footnotesize
    \begin{center}
        \begin{scriptsize}
        \begin{tabular}{r@{\hskip 1em}r@{\hskip 1em}l}
            \input{tables/sat/main/sat2016_main_alt_rankings}
        \end{tabular}
        \end{scriptsize}
        \caption{%
            Example rankings obtained using our proposed approach on the \emph{main} track from the 2016 SAT Competition.
            We show the solver groups in order, along with their respective fractional rank.
            In each group, solvers are ordered by the median of their competition scores in our bootstrap samples (official competition ranks are shown in parentheses).
            \vspace{-0.5cm}
        }
        \label{tab:exampleAltRankings}
    \end{center}
\end{table}

\section{IPC competition analysis}

Next, we consider three tracks from the deterministic part of the most recent international planning competition for which results are available, IPC 2014~\cite{ipc2014}. These tracks
are \emph{sequential agile, sequential satisficing} and \emph{sequential optimal},
which we abbreviate as ``seq-agl'', ``seq-sat'' and ``seq-opt''. We have selected these tracks due to the larger number of participants and the importance of the
deterministic tracks (especially the \emph{seq-sat} and \emph{seq-opt} tracks) in the planning community.
Table~\ref{tab:competitionSummaries} again shows the number of solvers and benchmark instances
considered in each of the tracks.
In each track, 14 sets of 20 problem instances were used, where each set was drawn from a distinct planning domain, for a total of 280 instances per track.
In order to preserve this benchmark set structure, which is a distinct feature of the planning competitions, we performed 
domain-stratified bootstrap sampling, by sampling with replacement within each domain, ensuring that 20 instances from each domain were included in every bootstrap sample.
As for the 2016 SAT Competition, we generated 10\,000 bootstrap replicates.

Figure~\ref{fig:ipc2014bootstrapCI} illustrates the 0.95 bootstrap confidence intervals for the top 10 solvers of the \emph{seq-sat} track, and
the IPC portion of Table~\ref{tab:empiricalFirstPlace} again shows multiple solvers ranked first in our samples (although here only four such solvers).
There are more obvious groupings of solvers here than for the 2016 SAT data, but we still observe many differences in ranks across the bootstrap replicates.
In the case of the \emph{seq-sat} track, there was a 2-way tie for first place between \emph{IBaCoP2} and \emph{IBaCoP}, a 4-way tie for group 2 and a 3-way tie for group 3. 
The empirical probabilities of ranking first are consistent with these results, with \emph{IBaCoP2} ranked first in 84.48\% of the samples, \emph{IBaCoP} ranked first in 15.08\% of the samples.
Compared to what we observed for the 2016 SAT Competition, the groups of statistically tied planners tend to be smaller
and the number of planners that have been ranked top in any bootstrap sample is very limited.
This is possibly due to the fact that in \emph{seq-sat}, planners have plenty of time (30 CPU minutes) for solving each
benchmark instance, and they are evaluated solely based on the quality of the best plan found within that time.

Even more extreme is the \emph{seq-opt} track (full results shown in Supplementary Material), where planners are run for 30 CPU
minutes and are scored based only on whether or not they find an optimal plan.
This track had clear first- and second-place solvers (\emph{SymBA*-2} and \emph{SymBA*-1}, respectively) in our robust ranking.
However, there was still a 3-way tie for group 3 and a 6-way tie for group 4.

Conversely, in the \emph{seq-agl} track,
planners are evaluated according to their running time, ignoring the quality of solutions.
In this track, our analysis confirms the clear winner, \emph{YAHSP3}, but
indicates a 6-way tie for group 2 and a 3-way tie for group 3.

We saw no rank inversions in planning results of the sort seen for the SAT competition,
but in an experiment performed with our per-domain bootstrap sampling disabled (so allowing different numbers of instances per domain)
we did detect several such inversions. Specifically, in the \emph{seq-agl} track, \emph{Freelunch} (competition rank 7) had fractional rank 3.5 (group 1) in our results,
while \emph{Fast Downward Cedalion} (competition rank 6) had a fractional rank of 8 (group 2). In the \emph{seq-opt} track, \emph{AllPACA} (competition rank 7)
had fractional rank 4.5 (group 3) while \emph{Dynamic-Gamer} (6) had a fractional rank of 9 (group 4).

\section{ImageNet ILSVRC Competition Analysis}

Finally, we consider 4 tracks of the most recent ImageNet ILSVRC competition, ILSVRC 2015.
These tracks are \emph{object classification+localization} (separately considering classification and localization accuracy),
\emph{object detection}, and \emph{object detection with video}. In each track, there was a version where competitors could only use
a provided training set ('provided data'), and a version with free choice of training set ('additional data').
Unlike for the SAT and AI planning competitions, in the case of the ILSVRC the 'problem instances' for which we have data
are actually entire image categories or classes. This means that individual competitors scores on these instances are much
more likely to be robust as they are already averages over many images.

However, as can be seen in Table~\ref{tab:competitionSummaries} and Figure~\ref{fig:imagenet2015bootstrapCI}, there is is still substantial
overlap in the performance of the competing solvers. With the exception of some of the 'additional data' tracks with very few competitors, we
see many tied solvers in our robust rankings, including in the top-10 in many cases.
Specifically, for the \emph{object classification+localization} track with provided data, when looking at classification accuracy ('class-prov'),
we found 15 ranking groups containing more than one solver; this included a 3-way tie for first place between \emph{MSRA 7}, \emph{MSRA 6} and \emph{ReCeption 6}.

The results for the other 'provided data' tracks are qualitatively similar. We note the presence of several inversions of solver
rank in each of the provided data tracks, with an interesting case in the \emph{object detection with video} track.
There, \emph{ITLab VID - Inha 11} (competition rank 4) had a fractional rank of 7 (group 3) in our robust rankings, placing it behind
\emph{UIUC-IFP 11}, \emph{12}, and \emph{13} (competition rank 6) which were in group 2. Additionally, \emph{RUC BDAI 11} (17) was placed in group 4, higher
than \emph{HiVision 12} (14), \emph{1-HKUST 12} (15) and \emph{1-HKUST 11} (16).

\section{Discussion}

We anticipate several questions interested readers may ask.
Firstly, what do our results mean for competitions in AI and beyond?
There appears to be something inherently unsatisfactory about declaring a tie between 17 solvers, as we do for top rank in the main track of the 2016 SAT Competition.
Of course, we find it even less satisfactory to declare a winner based on what can meaningfully be interpreted as statistical noise.
However, to stick with this example, we do not claim that 17 gold medals should have been awarded in this particular competition track, nor do we mean to imply that the official winner did not deserve the top place on the podium.
However, if we wish to draw conclusions from the competition results that go at least somewhat beyond the particular sample of benchmark instances used in this case, we should acknowledge that the performance differences between the official winner and the 16 next best solvers are insignificant in a precisely defined statistical sense.
This also helps us recognise qualitatively different situations, as encountered for the SAT 2016 \emph{agile} track
and ILSVRC 2015 \emph{object detection with provided data} track (among others),
where there were single, statistically well supported winners.
Furthermore, we feel that cases in which the ranking from the official results is inconsistent with that induced by the median scores across the bootstrap replicates or, worse, our robust rankings are inherently problematic, and that it is useful to know about cases involving highly-ranked solvers.

Secondly, what do our results mean for future competitions?
We believe that competitions are useful and that our robust ranking can provide meaningful guidance in addition to traditional scoring, not only for interpreting competition results but also for designing future competitions.
Large groups of solvers tied for top ranks can be avoided by using larger or differently constructed sets of benchmark instances (to a certain extent, this is seen in our ILSVRC results).
Robust rankings from past competitions can provide some guidance in this context without creating potentially unfair bias.
If statistical ties in competition results (in particular, podium ranks) are undesired, we believe that 
tie-breaking mechanisms should be used that minimise the risk of obtaining a ranking that is inconsistent with the robust ranking, such as median score over bootstrap replicates.

Thirdly, how much confidence can we have in our robust rankings?
While we ensure that the probability for type I errors within each group is bounded, 
we do not control for the compounding of error probability across the sequence of groupings we obtain. 
Therefore, our confidence that \emph{none} of the solvers in $\mathcal{S}$ is ranked lower than it should be decreases with the number of solvers considered.
Obviously, our procedure could be modified to avoid this effect, but we did not consider such a modification, partly in light of our observation that standard competition settings already produce large groups of statistically tied solvers, and bounding the error probability across all solvers would generate even more ties. Furthermore, and perhaps more importantly, for most competitions, only the top-ranked solvers generate substantial interest and impact, and therefore, the number of tied groups of interest does not increase with the size of $\mathcal{S}$.

Finally, what about type II errors, i.e., failure to reject an incorrect null-hypothesis that one solver performs no better than another? 
We note that in our robust ranking approach, the consequence of a type II error is that a solver is declared tied with another, although in reality it performs statistically worse. As a result, a solver might erroneously move up in (fractional) rank, stealing some thunder from solvers it now appears to be tied with. We find this much less problematic than type I errors, which move solvers down in (fractional) rank and may cost them a podium position.
Type II error probability decreases with growing sample size, here: number of problem instances used in a competition. This indicates, at least in principle, a way to reduce the probability of type II errors. 
At the same time, we believe that the same effect might be achievable by refining our approach further to increase the power of the statistical tests used to induce our robust ranking.

\section{Conclusions}


Competitions play an important role in AI.
Their outcomes routinely guide research and applications, but
can be affected by numerous factors.
Among these factors, the set of benchmark instances used in a competition
is of pivotal importance, and it is intuitively clear that competition results
cannot be expected to generalise to very different sets of benchmark instances.
However, they should be stable under very minor modifications or ideally across
instance sets sampled from the same underyling distribution.

In this work, we have quantified the 
effect on competition rankings under an approximation of such samples.
We developed an approach based on established concepts from resampling statistics
and illustrated the results it produces by applying it to three recent international competitions.

It should be noted that our approach is very general and can
be applied to existing competition results with negligible computational and human effort:
neither new solver runs nor new benchmark instances are required.
It allows us to identify meaningful performance differences and helps to recognise those that should be
considered statistically significant. Given this low overhead, we believe this should be
applied in competitions. At the very least, to call attention to
particularly noteworthy achievements, and for the scientific interpretation of
competition results.

Moreover, the proposed approach can be used, in combination with other
methods (e.g., that of \citeauthor{hoos2013bench} \citeyear{hoos2013bench}) to facilitate the
construction of benchmark sets;
e.g., it could be used to obtain benchmarks that minimise
performance ties of top-ranked solvers from previous competitions.
Our approach could also help to reduce the computational burden of competitions,
by running them in a racing-like fashion, where solvers are iteratively evaluated
on incrementally larger benchmark sets and eliminated as soon as their performance
is shown to fall statistically significantly below that of the current front runner.
We see the methodology introduced in this work as an
important step in improving the guidance provided by
competitions.


\vspace*{15mm}
\noindent {\bf Note to readers:} This paper has originally been written in late 2016, but due to unfortunate circumstances not been published until now. Upon request by several members of the AI research community, we are hereby making it available in a citable form, in the hope that others may be able to build on it.

\clearpage

\bibliographystyle{aaai}
\bibliography{references}

\end{document}

%% file: tables/single_instance_removal_summary.tex
                   &                    & \multicolumn{5}{c}{\# \textbf{Ranking Changes}} \\
                   &                    & \textbf{All}              & \multicolumn{2}{c}{\textbf{Top-10}}   & \multicolumn{2}{c}{\textbf{Top-3}} \\
\textbf{Track}     & \textbf{\# inst.}  &                           & comp.             & order             & comp.             & order \\
\midrule
main               &  500               & 164                       &  0                & 105               &  0                & 36 \\
main-app           &  300               &  93                        & 56                &  20               &  3                &  8 \\
main-craft         &  200               &  71                        &  6                &  26               &  7                &  4 \\
agile              & 5000               &  30                        &  0                &  30               & 20                &  0 \\
random             &  240               &   6                        &  0                &   6               &  0                &  4 \\
nolimit            &  329               &  66                        & 27                &  29               &  0                &  0 \\

\midrule
seq-agl            &  280               & 150                        &  0                & 136               & 62                &  0 \\
seq-sat            &  280               &  45                        &  0                &  45               &  0                &  0 \\
seq-opt            &  280               &  71                        &  0                &  25               &  0                &  0 \\

\midrule
loc-prov           & 1000               & 423                        &  0                &   0               &  0                &  0 \\
loc-add            & 1000               &   0                        &  0                &   0               &  0                &  0 \\
class-prov         & 1000               & 145                        &  0                &   2               &  0                &  1 \\
class-add          & 1000               &   0                        &  0                &   0               &  0                &  0 \\
det-prov           &  200               &  32                        &  0                &   0               &  0                &  0 \\
det-add            &  200               &   1                        &  0                &   1               &  0                &  0 \\
det-vid-prov       &   30               &  11                        &  0                &   0               &  0                &  0 \\
det-vid-add        &   30               &   0                        &  0                &   0               &  0                &  0 \\

%% file: tables/competition_summaries.tex
                   &                        &                          & \multicolumn{4}{c}{\textbf{All Solvers}}                                          & \multicolumn{4}{c}{\textbf{Top 10}}                                                   & \multicolumn{4}{c}{\textbf{Top 3}} \\
\cmidrule(r){4-7} \cmidrule(r){8-11} \cmidrule{12-15} \vspace{-0.4cm}\\
\textbf{Track}     & \textbf{Solvers}       & \textbf{Inst.}           & \textbf{Groups}       & \textbf{Ties}     & \textbf{Inv.}     & \textbf{Mean IQR} & \textbf{Groups}       & \textbf{Ties}     & \textbf{Inv.}     & \textbf{Mean IQR}     & \textbf{Groups}       & \textbf{Ties}     & \textbf{Inv.}     & \textbf{Mean IQR} \\
\midrule
main               & 29                     &  500                     &  4                    & 181               &  0                & 4.224             & 1                     & 45                & 0                 & 4.75                  & 1                     & 3                 & 0                 & 3.50 \\
main-app           & 29                     &  300                     &  3                    & 351               &  0                & 7.569             & 1                     & 45                & 0                 & 7.80                  & 1                     & 3                 & 0                 & 4.00 \\
main-craft         & 29                     &  200                     &  3                    & 135               &  1                & 3.241             & 1                     & 45                & 0                 & 3.20                  & 1                     & 3                 & 0                 & 1.50 \\
agile              & 26                     & 5000                     & 10                    & 30                &  1                & 0.923             & 5                     & 9                 & 1                 & 0.90                  & 2                     & 1                 & 0                 & 1.00 \\
random             &  9                     &  240                     &  3                    & 10                &  0                & 0.389             & 3                     & 10                & 0                 & 0.39                  & 1                     & 3                 & 0                 & 0.67 \\
nolimit            & 20                     &  329                     &  5                    & 42                &  0                & 2.025             & 3                     & 18                & 0                 & 2.30                  & 1                     & 3                 & 0                 & 0.67 \\

\midrule
seq-agl            & 15                     &  280                     & 5                     & 22                &  0                & 0.933             & 3                     & 18                & 0                 & 1.00                  & 2                     & 1                 & 0                 & 0.67 \\
seq-sat            & 20                     &  280                     & 9                     & 19                &  0                & 0.550             & 4                     & 10                & 0                 & 0.80                  & 2                     & 1                 & 0                 & 0.33 \\
seq-opt            & 17                     &  280                     & 7                     & 22                &  0                & 0.735             & 4                     & 13                & 0                 & 0.95                  & 3                     & 0                 & 0                 & 0.17 \\

\midrule
loc-prov           & 62                     & 1000                     & 35                    & 46                &  3                & 0.419             & 5                     &  7                & 0                 & 0.00                  & 2                     & 1                 & 0                 & 0.00 \\
loc-add            &  4                     & 1000                     &  4                    &  0                &  0                & 0.000             & 4                     &  0                & 0                 & 0.00                  & 3                     & 0                 & 0                 & 0.00 \\
class-prov         & 62                     & 1000                     & 28                    & 92                &  4                & 0.935             & 5                     &  7                & 0                 & 0.70                  & 1                     & 3                 & 0                 & 1.33 \\
class-add          &  4                     & 1000                     &  3                    &  1                &  0                & 0.000             & 3                     &  1                & 0                 & 0.00                  & 3                     & 0                 & 0                 & 0.00 \\
det-prov           & 39                     &  200                     & 21                    & 37                &  2                & 0.667             & 6                     &  7                & 0                 & 0.40                  & 3                     & 0                 & 0                 & 0.00 \\
det-add            & 13                     &  200                     & 12                    &  1                &  0                & 0.154             & 9                     &  1                & 0                 & 0.20                  & 3                     & 0                 & 0                 & 0.00 \\
det-vid-prov       & 23                     &   30                     &  8                    & 38                &  6                & 0.696             & 4                     & 13                & 3                 & 0.40                  & 2                     & 1                 & 0                 & 0.33 \\
det-vid-add        & 10                     &   30                     &  5                    &  6                &  0                & 0.300             & 5                     &  6                & 0                 & 0.30                  & 2                     & 1                 & 0                 & 0.00 \\

%% file: tables/empirical-first-place-truncated.tex
                                        & \textbf{Fraction} \\
\textbf{Solver}                         & \textbf{ranked 1\textsuperscript{st}} \\
\midrule
MapleCOMSPS\_DRUP seq\_LRB\_VSIDS\_drup & 38.16\% \\
Riss6 noPP\_DRAT                        & 29.66\% \\
Lingeling bbc main default              & 21.15\% \\
13 additional solvers                   & 23.01\% \\

\midrule
IBaCoP2                                 & 84.48\% \\
IBaCoP                                  & 15.08\% \\
Mercury                                 &  0.43\% \\
MIPlan                                  &  0.01\% \\

\midrule
MSRA 6                                  & 60.85\% \\
MSRA 7                                  & 60.85\% \\
ReCeption 6                             & 39.18\% \\

%% file: tables/sat/main/sat2016_main_alt_rankings.tex
\textbf{Group} & \textbf{Fract. Rank} & \textbf{Solvers} \\
\midrule
1    &  9   & MapleCOMSPS\_DRUP seq\_LRB\_VSIDS\_drup (1), \\
     &      & Riss6 noPP\_DRAT (2), Lingeling bbc main default (3), \\
     &      & glucose\_hack\_kiel\_newScript main\_new (4), \\
     &      & abcdSAT\_drup default (5), tc\_glucose default (5), \\
     &      & MapleCOMSPS\_LRB\_DRUP LRB\_VSIDS\_drup (7), \\
     &      & Riss6 default\_DRAT (7), tb\_glucose default (9), \\
     &      & Beans And Eggs default (9), glucose default (11), \\
     &      & COMiniSatPS Chandrasekhar DRUP drup (12), \\
     &      & GHackCOMSPS\_DRUP ghack\_drup (13), \\
     &      & MapleGlucose default (14), Glucose\_nbSat default (16), \\
     &      & glue\_alt default (16), Splatz 06v main default (16) \\
2    & 22.5 & glucosePLE default (18), \\
     &      & MapleCOMSPS\_CHB\_DRUP CHB\_VSIDS\_drup (19), \\
     &      & glueminisat-2.2.10-81-main maintrack (20), \\
     &      & MapleCMS default (21), CHBR\_glucose default (22), \\
     &      & Riss6 blackbox\_DRAT (22), CHBR\_glucose\_tuned default (25), \\
     &      & cmsat5\_autotune2 default (25), cmsat5\_main2 default (25), \\
     &      & gulch default (27) \\
3    & 28   & Scavel\_SAT default (28) \\
4    & 29   & YalSAT 03r default (29) \\